%% file: main.tex
\documentclass[accepted]{article}

\usepackage{balance}
\usepackage{fontawesome}
\usepackage[ruled,vlined,linesnumbered]{algorithm2e}
\usepackage{amsthm, amsmath, amsfonts, amssymb}
\usepackage{mathtools}

\usepackage[utf8]{inputenc} 
\usepackage[T1]{fontenc}    
\usepackage[backref=page]{hyperref}       
\usepackage{includes/icml2023}

\usepackage{nameref}
\usepackage{varioref}
\usepackage{hyperref}
\usepackage[capitalize,noabbrev]{cleveref}
\usepackage[font=small,labelfont=bf]{caption}
\usepackage[title]{appendix}
\usepackage{url}            
\usepackage{booktabs}       
\usepackage{amsfonts}       
\usepackage{nicefrac}       
\usepackage{microtype}      
\usepackage{xcolor}         
\usepackage{wrapfig}
\usepackage{graphicx}
\usepackage{subcaption}
\usepackage{mathtools}
\usepackage{amsmath,bm}
\usepackage{dsfont}
\usepackage{adjustbox}

\icmltitlerunning{Differentiable Multi-Target Causal Bayesian Experimental Design}

\definecolor{mydarkblue}{rgb}{0,0.08,0.45}
\hypersetup{
    colorlinks=True,
    linkcolor=mydarkblue,
    citecolor=mydarkblue,
    filecolor=mydarkblue,
    urlcolor=mydarkblue,
}

\include{macros}

\begin{document}

\twocolumn[
\icmltitle{Differentiable Multi-Target Causal Bayesian Experimental Design}


\icmlsetsymbol{equal}{*}
\icmlsetsymbol{equal1}{$\dagger$}

\begin{icmlauthorlist}
\icmlauthor{Yashas Annadani}{equal,kth,helmholtz}
\icmlauthor{Panagiotis Tigas}{equal,oatml}
\icmlauthor{Desi R. Ivanova}{stats}
\icmlauthor{Andrew Jesson}{oatml}
\icmlauthor{Yarin Gal}{oatml}
\icmlauthor{Adam Foster}{equal1,msr}
\icmlauthor{Stefan Bauer}{equal1,helmholtz,tum,cifar}

\end{icmlauthorlist}

\icmlaffiliation{kth}{KTH Stockholm, Sweden}
\icmlaffiliation{helmholtz}{Helmholtz AI}
\icmlaffiliation{oatml}{OATML, University of Oxford}
\icmlaffiliation{stats}{Department of Statistics, University of Oxford}
\icmlaffiliation{msr}{Microsoft Research}
\icmlaffiliation{cifar}{CIFAR Azrieli Global Scholar}
\icmlaffiliation{tum}{TU Munich}

\icmlcorrespondingauthor{Yashas Annadani}{yashas.annadani@gmail.com}
\icmlcorrespondingauthor{Panagiotis Tigas}{ptigas@robots.ox.ac.uk}
\icmlcode{\href{https://github.com/yannadani/DiffCBED}{https://github.com/yannadani/DiffCBED}}

\icmlkeywords{Machine Learning, ICML}

\vskip 0.3in
]



\printAffiliationsAndNotice{\icmlEqualContribution,\icmlEqualSupervision} 

\begin{abstract}

We introduce a gradient-based approach for the problem of Bayesian optimal experimental design to learn causal models in a batch setting --- a critical component for causal discovery from finite data where interventions can be costly or risky. Existing methods rely on greedy approximations to construct a batch of experiments while using black-box methods to optimize over a \emph{single target-state} pair to intervene with. In this work, we completely dispose of the black-box optimization techniques and greedy heuristics and instead propose a conceptually simple end-to-end gradient-based optimization procedure to acquire a set of optimal intervention target-state pairs. Such a procedure enables parameterization of the design space to efficiently optimize over a batch of \emph{multi-target-state} interventions, a setting which has hitherto not been explored due to its complexity.  We demonstrate that our proposed method outperforms baselines and existing acquisition strategies in both single-target and multi-target settings across a number of synthetic datasets.


\end{abstract}

\begin{figure}[!t]
\centering     
\label{fig:int-inf-des}
\includegraphics[width=0.44\textwidth]{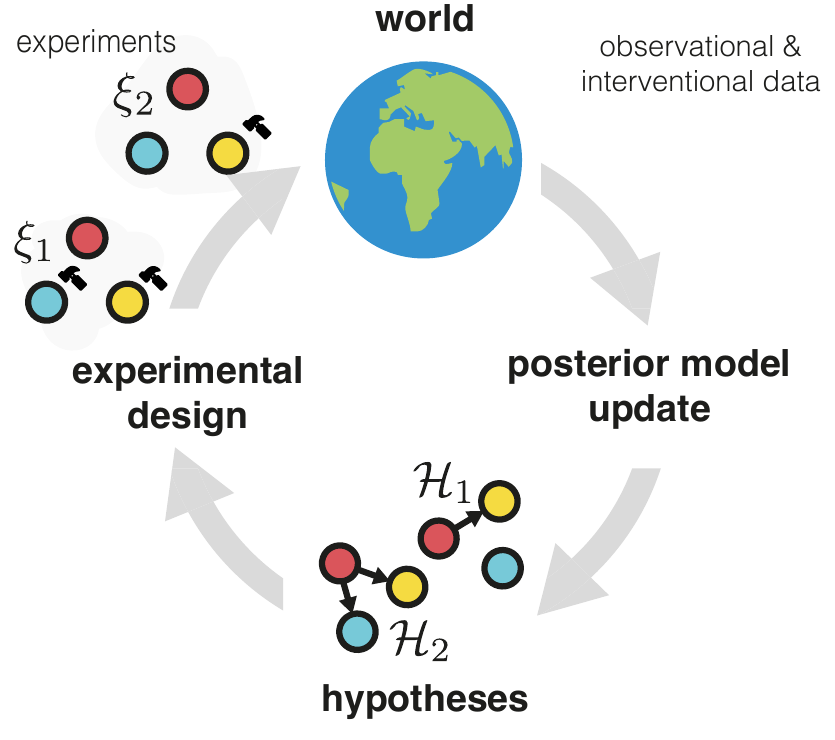}
\caption{Causal Bayesian Experimental Design optimizes experiments that help disambiguate between competing causal hypotheses.}
\end{figure}

\section{Introduction}
Imagine a scientist entering a wet lab to conduct experiments in order to discover the underlying causal relations within the system of interest. The scientist first comes up with some hypotheses, based on prior knowledge and past observations. Then, based on the formed hypotheses, an experimentation protocol to disambiguate between the competing hypotheses is devised. Additionally, because of the financial and ethical costs and risks involved in such experimentation, it is in the scientist's interest to minimize the number of batches required.
\begin{table*}[h!]
\centering
\caption{Comparison of different BOED for causal discovery methods based on their design space assumptions.}
\label{tab:assumptions}
\resizebox{42em}{!}{
\begin{tabular}{lccccc}
\hline
\textbf{}                        & \multicolumn{5}{c}{\textbf{Design Space Assumptions}}                                                                                                                                                                                                                                                                                                                                                                                                                          \\ \hline
\multicolumn{1}{l|}{}            & \multicolumn{1}{c|}{\begin{tabular}[c]{@{}c@{}}Target Acquisition \\ (Single Target)\end{tabular}} & \multicolumn{1}{c|}{\begin{tabular}[c]{@{}c@{}}State Acquisition\\ (Single Target)\end{tabular}} & \multicolumn{1}{c|}{\begin{tabular}[c]{@{}c@{}}Target Acquisition\\ (Multi-target)\end{tabular}} & \multicolumn{1}{c|}{\begin{tabular}[c]{@{}c@{}}State Acquisition\\ (Multi-target)\end{tabular}} & \begin{tabular}[c]{@{}c@{}}Batch \\ Acquisition\end{tabular} \\ \hline
\multicolumn{1}{l|}{\citet{murphy2001active}}      & \checkmark & ~ & ~ & ~ & ~\\
\multicolumn{1}{l|}{\citet{tong2001active}} & \checkmark & ~ & ~ & ~ & ~ \\
\multicolumn{1}{l|}{\citet{cho2016reconstructing}}         & \checkmark & ~ & ~ & ~ & ~ \\
\multicolumn{1}{l|}{\citet{agrawal2019abcd}}        & \checkmark & ~ & ~ & ~ & \checkmark \\
\multicolumn{1}{l|}{\citet{toth2022active}}        & \checkmark & \checkmark & ~ & ~ & ~ \\
\multicolumn{1}{l|}{\citet{tigas2022interventions}}    & \checkmark & \checkmark & ~ & ~ & \checkmark \\
\multicolumn{1}{l|}{\citet{sussex2021near}}        & \checkmark & ~ & \checkmark & ~ & \checkmark \\
\multicolumn{1}{l|}{\textbf{Ours}}        & \checkmark & \checkmark & \checkmark & \checkmark & \checkmark 
\end{tabular}
}
\end{table*}
This process is known as experimental design, and assuming that the question of interest concerns discovering the causal structure of the system of interest, the process is known as experimental design for causal discovery. A Bayesian framework for this process has been proposed in prior work \cite{tong2001active,murphy2001active,cho2016reconstructing,agrawal2019abcd,sussex2021near,tigas2022interventions,toth2022active} which typically consists of updating an approximate posterior with past experimental data and using this updated posterior to compute experiments that are maximally informative, as evaluated by expected information gain---the objective of interest in \textit{Bayesian Optimal Experimental Design} (BOED)~\cite{lindley1956measure, chaloner1995bayesian}.


The problem of Bayesian Optimal Experimental Design for Causal Discovery (BOECD) is hard; computing the Bayesian posterior over Structural Causal Models (SCM)---a common framework for capturing causal relationships---is intractable. More importantly, the estimation and optimization of batch BOED objectives are computationally challenging, which has resulted in heuristics like the greedy batch strategy~\citep{agrawal2019abcd,sussex2021near} and soft-top-$k$ batch strategy~\citep{tigas2022interventions}. Additionally, in causal discovery, one is interested not only in identifying the variable (target) to intervene \textit{on} but also the state to set the intervention \textit{to}, resulting in a design space which is a product space of discrete targets and continuous states, making experimental design even more challenging. ~\citet{tigas2022interventions} proposed to use Bayesian Optimization (BO) to optimize over the continuous state-space of the interventions and a soft-top-$k$ heuristic to select a batch.

In this work, we propose a method for estimating and optimizing the BOED objective in a differentiable end-to-end manner, alleviating the inefficiencies introduced by the heuristics of the batch selection but also the black-box optimization over the intervening states. Specifically, we introduce estimators of mutual information based on nested estimation~\citep{ryan2003estimating,myung2013,huan2014gradient,foster2019variational} and importance sampling and extend it to the problem of causal discovery where the optimization is over both discrete nodes and continuous states. 
We cast the problem of batch experiment selection as a policy optimization where the policy uses either the Gumbel-Softmax or relaxed Bernoulli distribution~\cite{jang2016categorical,maddison2016concrete} for single target and multi-target settings respectively. When combined with the straight-through gradient estimator~\cite{bengio2013estimating} to optimize over the targets and gradient ascent over corresponding states, we can explore the space of optimal designs efficiently with powerful optimizers~\citep{kingma2014adam}. Our proposed method requires very few assumptions about the causal model and can explore wide range of design settings as compared to prior work (see Table~\ref{tab:assumptions}), thus opening up possibilities of experimental design for causal discovery in a broader range of applications.

\section{Related Work}

\paragraph{Differentiable Bayesian Optimal Experimental Design.}

\citet{huan2014gradient,foster2019variational,foster2020unified, kleinegesse2020minebed,kleinegesse2021gradientbased} developed a unified framework for estimating expected information gain and optimizing the designs with gradient-based methods. 
More recently, \citet{ivanova2022efficient} applied the Gumbel-Softmax relaxation within gradient-based BOED for contextual optimization.
In~\citet{ivanova2021implicit,foster2021deep}, the authors introduced a policy-based method for performing adaptive experimentation. More recently, work like~\citet{blau2022optimizing,lim2022policy} used reinforcement learning to train policies for adaptive experimental design.

\paragraph{Experimental Design for Causal Discovery.} One of the earliest works of experimental design for causal discovery in a BOED setting was proposed by ~\citep{murphy2001active} and ~\citep{tong2001active} in the case of discrete variables for single target acquisition. Since then, a number of works have attempted to address this problem for continuous variables in both the BOED framework~\citep{agrawal2019abcd,von2019optimal,toth2022active,cho2016reconstructing} and other frameworks~\citep{kocaoglu2017cost, gamella2020active, eberhardt2012number,lindgren2018experimental,mokhtarian2022unified,ghassami2018budgeted,olko2022trust,scherrer2021learning}. In contrast to the setting studied in this paper, of particular note are the approaches for experimental design for causal discovery in a non-BOED setting in the presence of cycles~\citep{mokhtarian2022unified} and latent variables~\citep{kocaoglu2017experimental}. 

Closer to our BOED setting are the approaches of \cite{tigas2022interventions} and \cite{sussex2021near}. Specifically, in \citep{tigas2022interventions}, the authors introduce a method for selecting single target-state pair with stochastic batch acquisition while \cite{sussex2021near} introduce a method for selecting a batch of multi-target experiments with a greedy strategy, based on a gradient-based approximation to mutual information, without selecting the intervention state. Our presented method in contrast can acquire a batch of multi-target-state pairs.
 
\paragraph{Bayesian Causal Discovery.} Designing experiments involves approximating the posterior of causal models to estimate mutual information. Approximating posterior distributions of causal structures and SCMs is hard due to combinatorially growing graph space~\citep{heinze2018invariant}. There are several works which have proposed various methods of approximate inference/ posterior sampling~\citep{friedman2013data,annadani2021variational,lorch2021dibs,cundy2021bcd,deleu2022bayesian,nishikawa2022bayesian}, which could be used for our proposed design framework.

\section{Background}
\subsection{Causality}
\paragraph{Notation.} Let $\mathbf{V}=\{1,\dots,d\}$ be the vertex set of any Directed Acyclic Graph (DAG) $\mathbf{g}=(\mathbf{V},E)$ and $\mathbf{X_V} = \{ \mathrm{X}_1, \dots, \mathrm{X}_d \} \subseteq \mathcal{X}$ be the random variables of interest indexed by $\mathbf{V}$.  
\paragraph{Structural Causal Model.} To deal with questions related to modelling causal relations between variables of interest, we employ the framework of Structural Causal Models (SCM)~\citep{peters2017elements}. In many fields of empirical sciences like network inference in single cell genomics~\citep{greenfield2010dream4}, inferring protein-signalling networks~\citep{sachs2005causal} and medicine~\citep{shen2020challenges}, SCMs provide a framework to model the effects of interventions~\citep{pearl2009causality}-- experiments which perturb the state of a variable to a desired state, thereby allowing to study the mechanisms which affect the downstream variables (for example, CRISPR gene knockouts~\citep{meinshausen2016methods} in genomics). Under this framework, each variable $X_i$ has an associated \emph{structural equation}, and is assigned a value which is a deterministic function of its direct causes $X_{\text{pa}(i)}$ as well as an exogenous noise variable $\epsilon_i$ with a distribution $P_{\epsilon_i}$:
\begin{equation}\label{eq:scm}
X_i \coloneqq f_i(X_{\text{pa}(i)}, \epsilon_i) \,\,\,\, \forall i\in \mathbf{V}\nonumber  
\end{equation}
$f_i$'s are mechanisms that relate how the direct causes affect the variable $X_i$. If the structural assignments are assumed to be acyclic, these equations induce a DAG $\mathbf{g}=(\mathbf{V},E)$ whose vertices correspond to the variables and edges indicate direct causes. An intervention on any variable $X_i$ corresponds to changing the structural equation of that variable to the desired state (value), $X_i\coloneqq s_i$, where $s_i\in \mathcal{X}_i$. It is denoted by the $\mathrm{do}$-operator~\citep{pearl2009causality} as $\mathrm{do}(X_i=s_i)$.

In this work, we assume that the SCM is causally sufficient, i.e. all the variables are measurable, and the noise variables are mutually independent. Though the mechanisms $f$ can be nonparametric in the general case, we assume that there exists a parametric approximation to these mechanisms with parameters $\bgamma\in \Gamma$. In the case of linear SCMs, $\bgamma$ corresponds to the weights of the edges in $E$. We further require that the functions $f$ are differentiable with respect to their parameters. Many classes of SCMs fall under this category, including the most commonly studied one -- the Gaussian additive noise models (ANM)\footnote{Note that differentiability of $f$ is the only assumption we require with respect to an SCM. We do not require that the noise is additive. For clarity of exposition, we restrict our focus to an ANM as they are the most commonly studied class of SCMs.}: 
\begin{equation}\label{eq:scm_additive}
X_i \coloneqq f_i(X_{\text{pa}(i)}; \bgamma_i) + \epsilon_i,\,\,\,\epsilon_i\sim \mathcal{N}(0, \sigma^2_i) \nonumber
\end{equation}

\paragraph{Bayesian Causal Discovery.} If the SCM for a given set of variables $\mathbf{X_V}$ is unknown, it has to be estimated from a combination of observational data (data obtained in an unperturbed state of a system) and experimental data under an intervention. This problem is called causal induction or causal discovery~\citep{spirtes2000causation}. This amounts to learning the parameters of the unknown SCM given by DAG $\mathbf{g}$, parameters of mechanisms, $\bgamma= \begin{bmatrix*} \gamma_1, \dots, \gamma_d \end{bmatrix*}$, and variances, $\sigma^2= \begin{bmatrix*} \sigma^2_1, \dots, \sigma^2_d \end{bmatrix*}$. For notational brevity, henceforth we denote $\bphi = (\bgamma, \sigma^2)$ and all the parameters of interest with $\btheta=(\bg, \bphi)$. In Bayesian causal discovery~\citep{heckerman1997bayesian}, the parameters of SCM are treated as random variables whose beliefs are updated according to the Bayes rule. A Bayesian method for causal discovery is preferable to model epistemic uncertainty about the model due to finite data as well as characterize equivalence classes of SCM like Markov Equivalence Class (MEC) in the case of non-identifiability~\citep{peters2012identifiability}. Interventions improve identifiability, but they have to be planned carefully. After acquiring interventional data, Bayesian methods update the posterior distribution to reduce the uncertainty of the SCM.

\subsection{Bayesian Optimal Experimental Design} \textit{Bayesian Optimal Experimental Design} (BOED)~\citep{lindley1956measure, chaloner1995bayesian} is an information theoretic approach to the problem of selecting the optimal experiment to estimate any parameter $\btheta$. For BOED, the \emph{utility} of the experiment $\xi$ is  the mutual information (MI) between the observation $\mathbf{y}$ and $\btheta$:
{
\begin{align*}
\mathrm{U}_\text{BOED}(\xi) &\triangleq \mathcal{I}(\mathbf{Y}; \bTheta \mid \xi)  \\
&= \E_{p(\btheta)p(\mathbf{y} \mid \btheta, \xi)}[\log p(\mathbf{y} \mid \btheta, \xi)-\log p(\mathbf{y} \mid \xi) ]\nonumber
\end{align*}
}
This objective is also known as the \textit{Expected Information Gain} (EIG). The goal of BOED is to select the experiment that maximizes this objective ${\xi^* = \argmax_\xi \mathrm{U}_\text{BOED}(\xi)}$. Unfortunately, evaluating and optimizing this objective is challenging because of the nested expectations~\citep{rainforth2018nesting} and several estimators have been introduced~\citep{foster2019variational,kleinegesse2019efficient}, which can be combined with various optimization methods to select the designs~\citep{foster2020unified,ivanova2021implicit,foster2021deep,blau2022optimizing}.

A common setting, called \textit{static}, \textit{fixed} or \textit{batch} design, is to optimize $B$ designs $\{\xi_1,\dots,\xi_B\}$ at the same time. The designs are then executed, and the experimental outcomes are collected for a Bayesian update of the model parameters.

\subsection{Causal Bayesian Experimental Design}

\textit{Causal Bayesian Experimental Design} is concerned with designing the most informative experiments to identify the true SCM so that the number of experiments required is as few as possible. An experiment in causal discovery corresponds to picking the intervention targets $I\in\mathcal{P} (\mathbf{V})$ and the corresponding states $S^I\in \underset{i\in I}{\cup} \mathcal{X}_i$ to set those targets to. A key component of such methods is computing a posterior over the parameters of the SCM. However, computing the posterior is a difficult task since the number of DAGs grows exponentially in the number of variables. Nevertheless, a plethora of methods exist~\cite{friedman2013data,annadani2021variational,lorch2021dibs,cundy2021bcd} which can be used with our approach.

Having access to such posterior models, one can estimate the EIG objective. One difficulty that still remains though is that optimizing the EIG objective over the experiments is a mixed discrete and continuous optimization problem, for which previous work has proposed to find the optimal value per candidate target via the use of black-box methods like \textit{Bayesian Optimization} (BO)~\cite{tigas2022interventions}. Additionally, for the construction of the batch of experimental designs, a greedy approximation is used to incrementally select experiments, a method that is $1-\frac{1}{\epsilon}$-approximate to the optimal solution \citep{krause2014submodular}.

\section{Differentiable Causal Bayesian Experimental Design}

Let $\bTheta$ be a random variable that models the uncertainty in the parameters of the true SCM, of which $\btheta\coloneqq(\mathbf{g},\bphi)$ is a realization. An experiment to identify an intervention is denoted by $\xi \coloneqq \design \coloneqq \mathrm{do}(\mathbf{X}_I = S^I)$, where $I\in\mathcal{P}(\mathbf{V})$ is a set of target indices in the multi-target setting, and $S^I$ are the corresponding states of those targets under intervention. The outcome of the experiment is denoted by $\mathbf{y} \sim P\left(\mathrm{X}_1 = \mathrm{x}_1, \dots, \mathrm{X}_d = \mathrm{x}_d \mid \mathrm{do}\left(\mathbf{X}_I = S^I \right)\right) = p(\mathbf{y} \mid \xi)$. 
Here, $\mathbf{y}$ is an instance of the random variable $\mathbf{Y} \subseteq \mathcal{X}$ distributed according to the interventional distribution\footnote{Note that when $I=\varnothing$, it corresponds to an observational/ non-experimental setting. In this case, $\mathbf{Y} = \mathbf{X_V}$.}. Due to causal sufficiency, the likelihood of data for a given $\btheta$ satisfies the causal Markov condition:
{
\begin{equation}\label{eq:scm_likelihood}
    p(\mathbf{y}\mid \btheta,\xi)=\prod_{j\in\mathbf{V}\setminus I}p\left(x_j|\bphi_j, \mathbf{x}_{\text{pa}_{\mathbf{g}}(j)},\mathrm{do}\left(\mathbf{X}_I = S^I \right)\right)
\end{equation}
}
Along with a prior $p(\btheta)$, the above equation defines a generative model of the data.

\paragraph{Design setting.} As in prior work~\citep{tigas2022interventions, sussex2021near}, we are interested in the setting of batch design where we design $B$ experiments at once before collecting experimental data. In other words, we seek a multiset of intervention targets and corresponding states which are jointly maximally informative about the parameters. We denote this multiset as $\xi_{1:B}\coloneqq(I_{1:B},S^{I}_{1:B})$. After executing a batch of experiments and collecting experimental outcomes, an experimenter might wish to design a new batch of experiments based on collected data (as summarized by the posterior distribution). Let $h_t$ denote experimental history $(\xi^1,\mathbf{y}^1),\dots,(\xi^t,\mathbf{y}^t)$ after $t$ batches of acquisition.  The BOED objective for this batch setting at any point $t$ is given by the joint mutual information:
{
\begin{equation}\label{eq:eig}
\begin{split}
    \mathcal{I}(\mathbf{Y}_{1:B}^t;& \bTheta \mid \xi_{1:B}^t, h_{t-1}) \\
=& \E_{\substack{p(\btheta\mid h_{t-1})\\p(\mathbf{y}_{1:B}^t \mid \btheta, \xi_{1:B}^t)}}\left[\log \frac{p(\mathbf{y}_{1:B}^t \mid \btheta, \xi_{1:B}^t)}{p(\mathbf{y}_{1:B}^t \mid \xi_{1:B}^t,h_{t-1})}\right]
\end{split}
\end{equation}
}%
where $\mathbf{Y}_{1:B}^t$ are the random variables corresponding to experimental outcomes for iteration $t$, $\mathbf{y_{1:B}^t}$ are the instances of these random variables and $\xi_{1:B}^t$ is the corresponding multiset of experimental designs. We drop the superscript $t$ from these variables for simplicity of exposition. Ideally, we wish to maximize the above objective by obtaining the gradients $\nabla_{\xi_{1:B}} \mathrm{I}$ and performing gradient ascent. However, the above objective is doubly intractable~\citep{rainforth2018nesting} and approximations are required. This usually leads to a two-stage procedure where the above objective is first estimated with respect to an inference network and then maximized with respect to designs~\citep{foster2019variational}, which can be typically inefficient~\citep{foster2020unified}.  

\subsection{Estimators of the Joint Mutual Information}

\subsubsection*{Nested Monte Carlo}\label{sec:nmc}

Following \cite{huan2014gradient,foster2020unified, foster2021deep}, we consider an estimator that allows for approximating the EIG objective while \textit{simultaneously} optimizing for the experiment $\xi$ that maximizes the objective via gradient-based methods. This estimator, called Nested Monte Carlo (NMC), is based on contrastive estimation of the experimental likelihood and has been extensively used in Bayesian experimental design \citep{ryan2003estimating,myung2013}. More precisely, assuming some past observational and interventional data $h_{t-1} = \{(\xi^1,\mathbf{y}^1),\dots,(\xi^{t-1},\mathbf{y}^{t-1})\}$, for every parameter sample from the posterior distribution $\btheta_0\sim p(\btheta\mid h_{t-1})$, a set of contrastive samples $\btheta_{1:L}\sim p(\btheta\mid h_{t-1})$ are considered to obtain a unified objective:
{\small
\begin{align} \label{eq:nmc} 
    \mathcal{U}_\text{NMC}^t(\xi_{1:B}) = 
    \E_{\substack{p(\btheta_{0:L} \mid h_{t-1})\\p(\mathbf{y}_{1:B} \mid \btheta_0, \xi_{1:B}))}} \left[\log \frac{p(\mathbf{y}_{1:B} \mid \xi_{1:B}, \btheta_0)}{
    \frac{1}{L} \sum_{\ell=1}^L p(\mathbf{y}_{1:B} \mid \xi_{1:B}, \btheta_\ell) )} \right]
\end{align}
}%
This estimator converges to the true mutual information as $L\to\infty$ \citep{rainforth2018nesting}.
If the design space is continuous, the optimal \emph{batch} of experiment $\xi^*_{1:B}$ can be found by \emph{directly} maximizing the NMC objective ($\xi^*_{1:B} \leftarrow \argmax_{\xi_{1:B}} \mathcal{U}_{\text{NMC}}^t(\xi_{1:B})$) with gradient-based techniques \citep{huan2014gradient}. 

The above objective requires estimating the posterior distribution $p(\btheta\mid h_{t-1})$ after every acquisition. For causal models, while it is generally hard to estimate this posterior due to DAG space of causal structures being discrete and super-exponential in the number of variables~\citep{tong2001active}, many approaches exist in the literature ~\citep{agrawal2019abcd,lorch2021dibs,cundy2021bcd}. These approximate posteriors can be nevertheless used for estimating the NMC objective.

\subsubsection*{Importance Weighted Nested Monte Carlo}\label{sec:iwnmc}
To establish an alternative path to estimating the mutual information, we begin by utilizing an observation from \citet{foster2019variational} that it is possible to draw the contrastive samples from a distribution other than $p(\btheta\mid h_{t-1})$ and obtain an asymptotically exact estimator, up to a constant $C$ that does not depend on $\xi^t_{1:B}$. Drawing samples from the \emph{original} prior $p(\btheta)$ gives the estimator
{
\begin{multline*}
     \mathcal{I}(\mathbf{Y}_{1:B}^t; \bTheta \mid \xi_{1:B}^t, h_{t-1}) - C =  \lim_{L \to \infty} \\
     \E_{\substack{p(\btheta_{0} | h_{t-1})p(\btheta_{1:L})\\p(\mathbf{y}_{1:B} | \btheta_0, \xi_{1:B})}} \left[\log \frac{p(\mathbf{y}_{1:B} | \xi_{1:B}, \btheta_0)}{
    \frac{1}{L} \sum_{\ell=1}^L p(\mathbf{y}_{1:B} | \xi_{1:B}, \btheta_\ell)p(h_{t-1} | \btheta_\ell )} \right]
\end{multline*}
}
The remaining wrinkle is that we must sample $\btheta_0$ from $p(\btheta_{0} | h_{t-1})$.
We propose the conceptually simplest approach of applying self-normalized importance sampling (SNIS) to the outer expectation.
The resulting objective, based on efficiently re-using samples in a leave-one-out manner, can optimize designs by just sampling parameters from the prior, without having to estimate the posterior:
{
\begin{multline}\label{eq:iwnmc}
     \mathcal{U}_{\text{IWNMC}}^t(\xi_{1:B}) = \\
     \E\left[ \sum_{m=1}^L \omega_m \log \frac{p(\mathbf{y}_{m,1:B}|\btheta_m,\xi_{1:B})}{\frac{1}{L-1}\sum\limits_{\ell\ne m} p(\mathbf{y}_{m,1:B}|\btheta_\ell,\xi_{1:B})p(h_{t-1}|\btheta_\ell)} \right]\raisetag{20pt}
\end{multline}
}
where $\btheta_{1:L}\sim p(\btheta_{1:L})$ are sampled from the original prior, $\mathbf{y}_{m,1:B} \sim p(\mathbf{y}_{1:B}|\btheta_m,\xi_{1:B})$ are all the experimental outcomes in the batch for parameter $\btheta_m$ 
and $\omega_m \propto p(h_{t-1}|\btheta_m)$ are self-normalized weights.

A full derivation is given in \Cref{sec:app:iwnmc}. 

 As IWNMC does not require any posterior estimation but instead relies entirely on the prior, it completely sidesteps the causal discovery process for designing experiments. This is a paradigm change from the NMC estimator which requires causal discovery through the estimation of the posterior. 
 
 However, we note that using IWNMC with just the prior (Eq.~\ref{eq:iwnmc}) as opposed to NMC (Eq.~\ref{eq:nmc}) comes with trade-offs. IWNMC typically requires a large $L$ to get a good estimate of the EIG. In high dimensions, this can be computationally infeasible. Having a small $L$ on the other hand might result in a failure case if the effective sample size of importance samples becomes 1. We can alleviate this issue if there is some prior information available which could be leveraged to design better proposal distributions. This might consist of knowledge of certain causal mechanisms of the system under study or access to some initial observational data. In such a case, a proposal distribution which encodes this information (for example with support on graphs which are in the Markov Equivalence Class (MEC) of the observational distribution) can be used instead of the prior. If no prior information is available or a good approximate inference technique is at our disposal, NMC is preferable in high dimensions. Surprisingly, we get good results on variables of size up to $5$ with IWNMC from just the prior and up to $40$ variables from a proposal distribution which has support on the MEC of observational distribution (see Sec~\ref{sec:results}).

\subsection{Optimizing over Targets and States (\texttt{DiffCBED})}
While the NMC estimator provides a unified objective to directly optimize over the designs $\xi_{1:B}$, it requires that the design space is continuous so that the gradients $\frac{\partial \mathcal{U}_\text{NMC}}{\partial I_{1:B}}$ and $\frac{\partial \mathcal{U}_\text{NMC}}{\partial S^I_{1:B}}$ can be computed. However, in the case of designing experiments for causal models, the challenge still remains that optimizing over intervention targets $I$ with gradient-based techniques is not possible because it is a discrete choice.

In order to address this problem, we introduce a \emph{design policy} $\pi_\phi$ with learnable parameters $\phi$ that parameterize a joint distribution over possible intervention targets and corresponding states. Instead of seeking the gradients $\frac{\partial \mathcal{U}_\text{NMC}}{\partial I_{1:B}}$ and $\frac{\partial \mathcal{U}_\text{NMC}}{\partial S^I_{1:B}}$, the goal now instead is to estimate $\frac{\mathcal{U}_\text{NMC}}{\partial \phi}$ so that policy can be updated to be close to optimal. Such a characterization of the design space allows us to use continuous relaxations of discrete distributions~\citep{maddison2016concrete,jang2016categorical} to obtain samples of designs and estimate NMC gradients.

 Let $\mathbf{I}$ and $\mathbf{S}$ be the random variables which model all possible intervention target combinations and states for a batch design respectively. While there are many possibilities of instantiating the policy in practice, we consider the simplest case where $\pi_\phi(\mathbf{I},\mathbf{S})\triangleq \pi_{\phi_n}(\mathbf{I})\pi_{\phi_m}(\mathbf{S})$. As the state space is continuous\footnote{If the state space is discrete, optimizing $\pi_{\phi_m}$ would be similar to $\pi_{\phi_n}$ which involves reparameterized gradients.}, $\pi_{\phi_m}$ can be either deterministic (a delta Dirac with $\phi_m \in \mathbb{R}^{B\times d}$) or Gaussian with $\phi_m \in \mathbb{R}^{2\times B\times d}$ parameterizing its mean and log variance. In this work, we found it sufficient to use a deterministic policy over the state space. For the interventional targets, ${\phi_n}\in \mathbb{R}^{B\times d}$ parameterizes the logits of different relaxed versions of discrete distributions depending on the setting, which we describe below. 
\begin{figure}[!b]
\centering     
\includegraphics[width=0.45\textwidth]{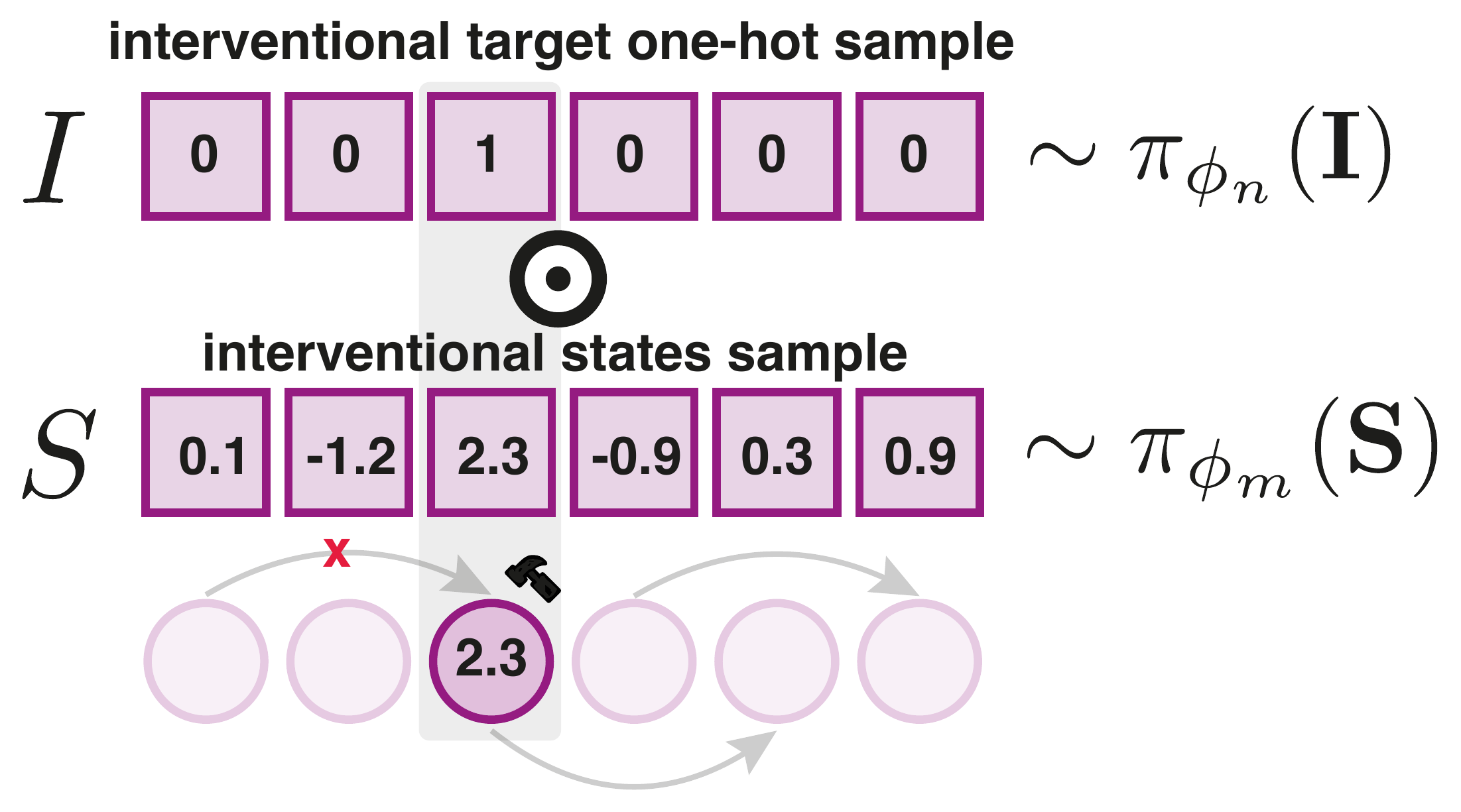}
\caption{ A design sample is obtained by first sampling $I_{1:B}\sim \pi_{\phi_n}(\mathbf{I})$, $S_{1:B}\sim \pi_{\phi_m}(\mathbf{S})$ and then setting states to be $S^I_{1:B}=S_{1:B}\odot I_{1:B}$. To obtain hard samples of $I$, we use the straight-through estimator~\citep{bengio2013estimating}. Illustration for $B=1$.}
\label{fig:policy}
\end{figure}


\begin{algorithm}[h]
  \SetEndCharOfAlgoLine{}
  \SetKwComment{Comment}{$\triangleright~$}{}
  \SetKwInOut{Input}{Input}
  \SetKwInOut{Output}{Output}

  \Input{$\mathcal{E}$ SCM Environment, $N$ Initial observational samples, $B$ Batch Size
 }
  $\mathcal{D}_{\text{obs}} \leftarrow \mathcal{E}.\text{sample}(N)$, $\mathcal{D}_{\text{int}} \leftarrow \varnothing $\\
  Train $q(\bTheta \mid \mathcal{D}_{\text{obs}})\approx p(\bTheta \mid \mathcal{D}_{\text{int}})$ using appropriate algorithm.\\
  \For{batch $t=1 \ldots $ $T$ Batches}{
    Initialize design policy parameters $\phi=\{\phi_n,\phi_m\}$: trainable logits $\phi_n$ for the targets; trainable parameters $\phi_m$ for the states.\\
    \For{ update step $c=1 \ldots C$}{
        \Comment{Sample Interventional Targets and States}
        $\{\xi_{1:B}^{(o)}\}_{o=1}^{O} \sim \pi_{\phi}(\mathbf{I}, \mathbf{S})$\\
        \Comment{Gradient ascent with straight-through gradient estimator}
        Update $\phi \rightarrow \phi + \alpha \frac{\partial}{\partial \phi} \frac{1}{O}\sum_{o=1}^{O}\left[ \mathcal{U}^t_{\text{NMC}}(\xi_{1:B}^{(o)}) \right]$
    }
    \Comment{Intervene with learned policy}
    $\xi_{1:B} \sim \pi_\phi$\\
    $\mathcal{D}_{\text{int}} \leftarrow \mathcal{D}_{\text{int}} \cup \mathcal{E}\text{.intervene}(\xi_{1:B})$\\
    Update the posterior $q(\bTheta \mid \mathcal{D}_{\text{obs}} \cup \mathcal{D}_{\text{int}})$.
  }
  \caption{Differentiable CBED (\texttt{DiffCBED})}
  \label{algo:method}
\end{algorithm}

\begin{figure*}[ht]
  \centering
  \includegraphics[width=0.9\textwidth]{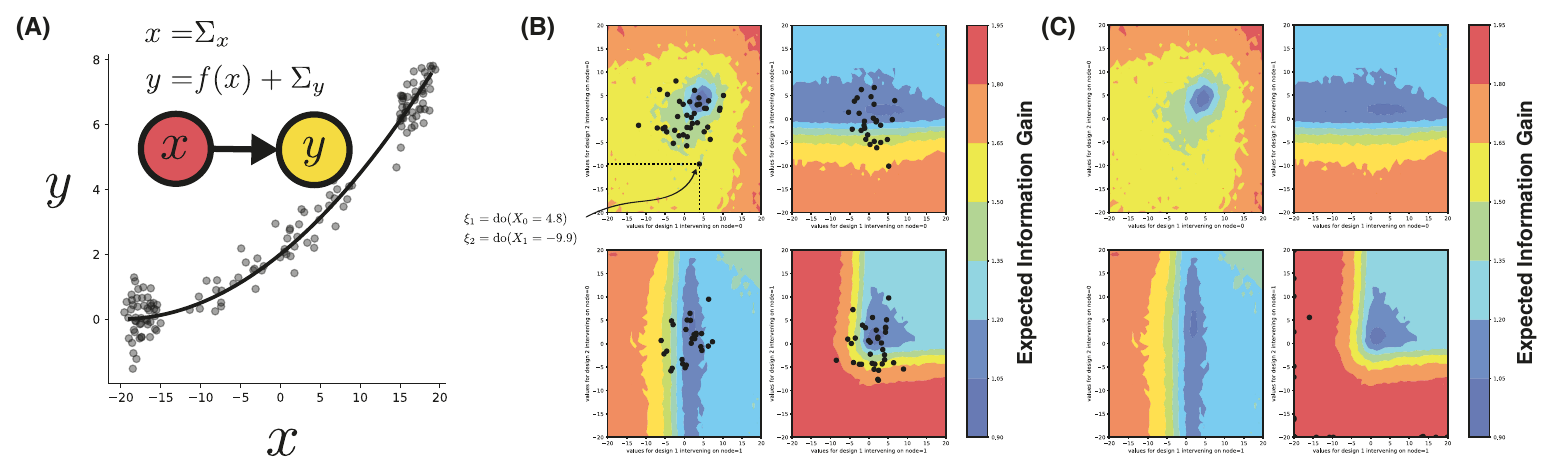}    
  \caption{Two variables and two experiments scenario. We assume a ground-truth graph $G_\text{T}$ of two nodes $X_1 \rightarrow X_2$. (A) The conditional distribution $p(X_2 \mid X_1)$ is shown in \textbf{(A)}. The corresponding SCM (Fig. A) is $x_1 = \Sigma_{x_1}$ and $x_2 = f(x_1) + \Sigma_{x_2}$. The four panels represent the EIG of all possible experiments of batch size two, when intervening on nodes $(0, 0), (0, 1), (1, 0), (1, 1)$. (B, C) Each panel shows how the EIG change on different interventional states. E.g. right top panel shows how EIG changes when applying interventions with states in ranges $[-20, 20]$. Fig. B Shows the designs before optimizing the objective and Fig. C after. As we can observe that the algorithm successfully places the designs (samples from the policy) on the high EIG (1.95) area of the plot ($\bullet$ on the plot).}
  \label{fig:didactic-scenario}  
\end{figure*}
\subsubsection*{Single Target ($q=1$)}

In this setting, the intervention targets are one-hot vectors, as demonstrated in \cref{fig:policy}. To sample one-hot vectors in a differentiable manner, we parametrize $\pi_{\phi_n}$ as a Gumbel-Softmax distribution~\citep{maddison2016concrete,jang2016categorical} over intervention targets, which is a continuous relaxation of the categorical distribution (in \emph{one-hot} form). Additionally, we use the straight-through (ST) gradient estimator~\cite{bengio2013estimating}.

\subsubsection*{Unconstrained Multi-Target ($q\leq d$)}

If instead of a continuous relaxation of the categorical distribution, we parametrise the policy $\pi_{\phi_n}$ as a continuous relaxation of the Bernoulli distribution (Binary Concrete)~\citep{maddison2016concrete}, we can now sample multi-target experiments. Since each interventional target sample will have at most $d$ non-zero entries, this policy is suitable for multi-target experiments with an unconstrained number of interventions per experiment. 

\begin{figure}[h!]
\centering     
\includegraphics[width=0.5\textwidth]{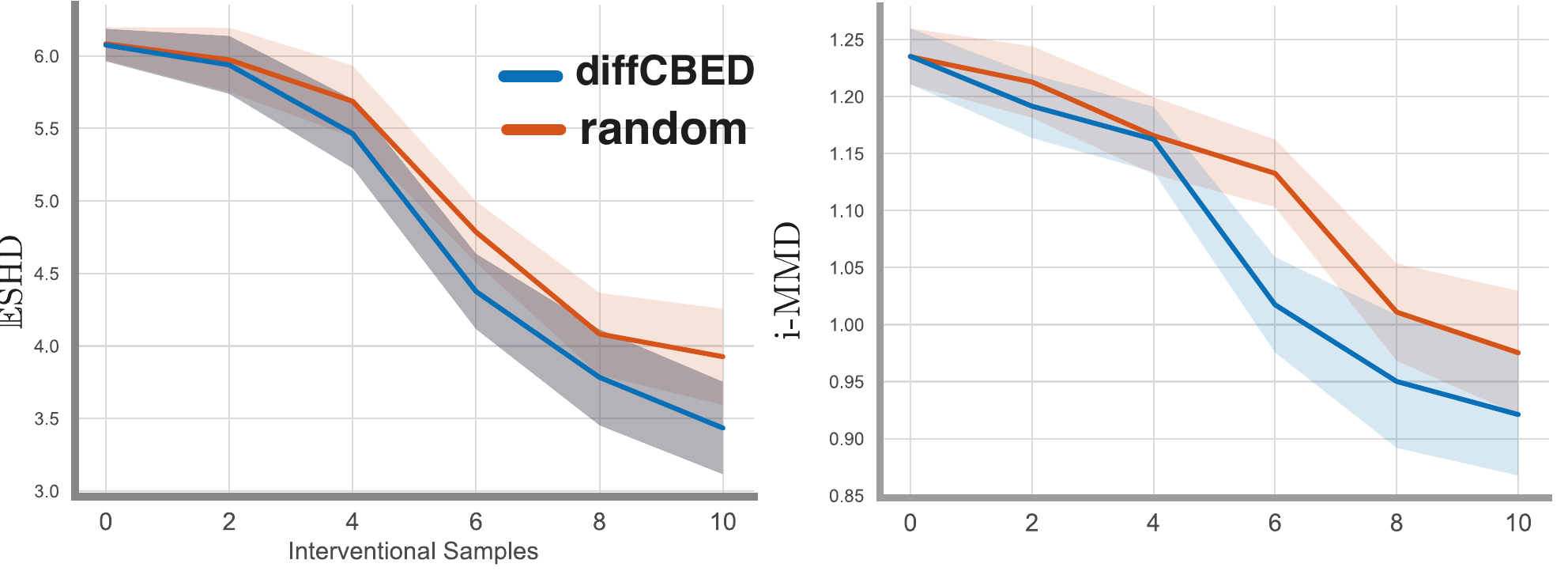}
\caption{We test the designs acquired with IWNMC estimator with just the prior as opposed to the random policy (with random target and state acquisition)  on variables of $5$ dimensions. Plots correspond to unconstrained multi-target setting with $B=2$ (shaded area represents $95\%$ confidence intervals - 60 seeds).}
\label{fig:policy_iwnmc_prior}
\end{figure}

\subsubsection*{Constrained Multi-Target ($q=k$)}

Finally, when considering a setting where the number of targets per intervention is exactly $k$. However, this is a significantly more challenging case, since the policy needs to select a subset of $k$ from $d$ nodes. By using a continuous relaxation of subset sampling, as introduced in ~\citet{xie2019reparameterizable}, combined with straight-through gradient estimator, we can efficiently optimize the policy to select a subset of nodes to intervene on. Note that with $k=1$, this would be equal to the single-target setting.

The \texttt{DiffCBED} algorithm is outlined in \cref{algo:method}.
\section{Experiments}
We evaluate the performance of our method on synthetic and semi-synthetic datasets and a range of baselines. We aim to investigate the following aspects empirically: (1) To what extent can we design good experiments without performing intermediate causal discovery/ posterior estimation with IWNMC estimator from the prior? (2) Ability to design good experiments with a proposal distribution with IWNMC (3) the performance of our policy-based design in combination with the differentiable NMC estimator in single-target and multi-target settings, as compared to suitable baselines.
\subsection{Bivariate Setting}
First, we demonstrate the method in a two variable setting to qualitatively assess the objective and the optimization method. Since computing the posterior over graphs and parameters is intractable in the general case, as a first step to study how well we can optimize the EIG objective, we assume a simple two-variable SCM. To compute the posterior we enumerate all DAGs with two nodes and parameterize the conditional distributions as neural network parameterized Gaussian distribution ($\mathcal{N}( X_i; \mu_\text{NN}(X_{\text{pa}(i)}), \sigma_\text{NN}(X_{\text{pa}(i)}))$). We compute the posterior over the parameters of the conditional distributions via Monte-Carlo dropout \cite{gal2016dropout}. We parameterize the intervention targets policy with Gumbel-Softmax and interventional states policy with a Gaussian distribution. The final policy consists of the logits of the Gumbel-Softmax and the sufficient statistics of the Gaussian distribution. We use Adam optimizer~\cite{kingma2014adam} to optimize the parameters of the policy. As we can see in Fig.\ref{fig:didactic-scenario}(B-C), the optimizer successfully concentrates the policy on the nodes and states that maximize the EIG objective.
\subsection{Results}
We present experimental results in various settings for the following metrics:
\subsubsection*{Evaluation metrics}
\noindent{\textbf{Expected SHD}}: This metric evaluates the expected structural hamming distance between the graphs sampled from the posterior model and the ground truth graph.

\noindent{\textbf{Expected F1-score}}: This metric evaluates the expected f1-score of each edge being present or absent in the graphs sampled from the posterior as compared to the ground truth graph.

\begin{figure*}[ht]
  \centering
  \includegraphics[width=0.9\textwidth]{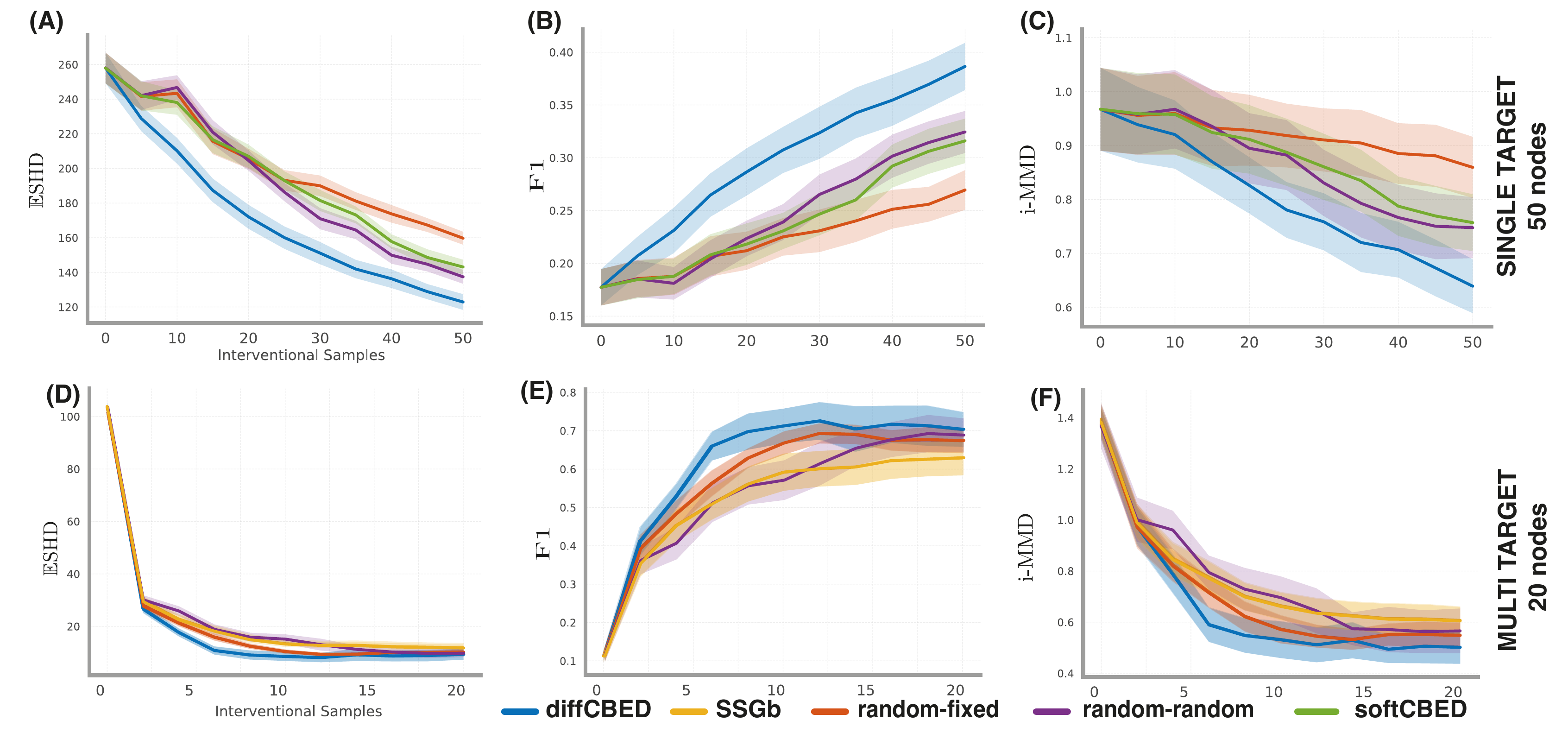} 
  \caption{\textbf{(A,B,C)} Single target-state design setting results for \erdos graphs with $d=50$ variables. \textbf{(D,E,F)} Multi target-state design setting results for \erdos graphs with $d=20$ variables. Each experiment was run with 30 random seeds (shaded area represents 95\% CIs)}
  \label{fig:synthetic_results}  
\end{figure*}

\noindent{\textbf{i-MMD}}: interventional MMD distance uses the non-parametric distance metric MMD~\cite{gretton2012kernel}. As compared to the graph evaluation metrics, this metric evaluates the interventional distributions induced by both the graph structure and the corresponding conditional distributions. We provide the full definition in Appendix~\ref{app:metrics}.

\subsection{Evaluation of the IWNMC estimator}
In this section, we consider optimizing the designs with respect to the IWNMC estimator entirely from the prior, introduced in~\ref{sec:iwnmc}, sidestepping the causal discovery procedure. As noted before, estimating posteriors of causal models is hard, so it is important to understand to what extent IWNMC can be considered a suitable candidate for designing good experiments in the absence of a posterior. For this setting, we sample from the prior distribution over graphs by first sampling an ordering of nodes at random and then sampling edges with probability $p=0.25$ which adhere to this topological order. We sample the mechanism parameters and noise variances of ANM at random from a Gaussian distribution with mean 0 and variance 1. 

Figure~\ref{fig:policy_iwnmc_prior} demonstrates results for $5$ variable unconstrained multi-target setting with batch size 2. For evaluation, we train DAG Bootstrap~\citep{friedman2013data} with GIES~\citep{hauser2012characterization} on the data acquired from each policy. We can see that we can recover the ground truth SCM faster than a random strategy. This is a surprising, but positive result given that our policy was trained entirely from samples from the prior. We also tested this approach for $10$ variables (results in Appendix~\ref{sec:IWNMC_full_results}). While this resulted in better performance of the policy as opposed to random in terms of downstream metrics, we observed effective sample size reach $1$, indicating that for $10$ dimensions or higher, we might need a better proposal distribution or a posterior estimate. 

\subsection{Baselines}
 Before we evaluate the IWNMC estimator with a proposal distribution more informative than the prior and the NMC estimator with a posterior estimate of SCM, we present the baselines with which we can compare the designs. 
\subsubsection*{Single target}

\noindent{\textbf{Random-Fixed}}: Uniform random selection of target, fixing the state to a value of $0$ (as introduced in \cite{agrawal2019abcd,tigas2022interventions}). \noindent{\textbf{Random-Random}}: Uniform selection of node, uniform selection of state (introduced in \cite{toth2022active}). \noindent{\textbf{SoftCBED}}: A stochastic approximation of greedy batch selection as introduced in \cite{tigas2022interventions}.

\subsubsection*{Multi-Target}

\noindent{\textbf{Random-Random}}: Multi-target version of uniform selection of node, uniform selection of value (introduced in \cite{toth2022active}). \noindent{\textbf{Random-Fixed}}: Multitarget version of uniform selection of node, fixed value to $5$ \cite{sussex2021near}, as suggested by the authors. \noindent{\textbf{SSGb}}: Finite sample baseline from~\cite{sussex2021near} with fixed value equal $5$. We emphasize that in contrast to our method, the baselines cannot select states, but they either assume a fixed predefined state or select a state at random.

\subsection{Evaluation in Higher Dimensions}\label{sec:results}
\subsubsection*{Evaluation of IWNMC with Proposal Distribution}\label{exp:constrained}
In this experiment, we consider $40$ variables, constrained $(q=5)$ multi-target and batch size $B=2$. Further, we use the same setup as~\citet{sussex2021near} to make a fair comparison as well as to construct a proposal distribution. For the proposal distribution, we use $800$ observational samples to train DAG Bootstrap~\cite{friedman2013data,agrawal2019abcd} and augment our posterior samples with samples of DAGs from the MEC of the true graph, to make sure that there is sufficient support within the MEC of the true graph (see~\citet{sussex2021near} for details). We then acquire a single batch of experiments from IWNMC estimator for our approach. For the baseline, we acquire a single batch of experiments from the estimator defined in \citep{sussex2021near}.
\begin{figure*}[h]
  \centering
  \includegraphics[width=.9\textwidth]{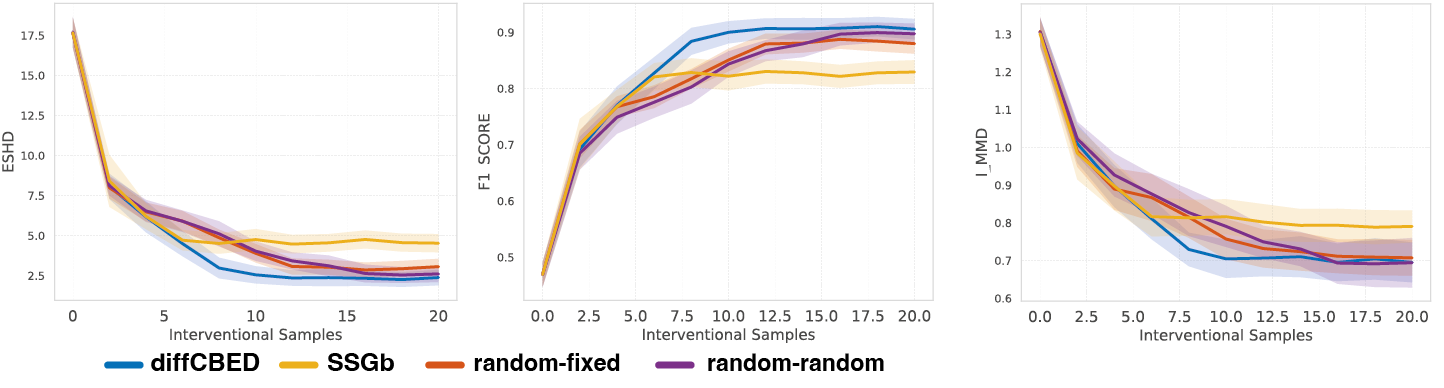} 
  \caption{Results on the E. Coli gene interaction network $(d=10)$ of the DREAM~\citep{greenfield2010dream4} simulator. The data is generated based on this graph by simulating the noise and mechanisms. We find that our method performs better than the baselines. Shaded area represents 95\% CI.}
  \label{fig:semi_synthetic_results}  
\end{figure*}

For random and SSGb baseline, we set the interventional state (value) to 5, as explained in \cite{sussex2021near}. Our approach doesn't fix the state to 5 but optimizes over states to perform the intervention with. In \cref{table:importance_weighted} we summarize our results. As we can see, our method outperforms random and SSGb by a great margin, indicating that with a good proposal distribution, IWNMC can still be a promising candidate in higher dimensions.
\subsubsection*{Results with NMC estimator}
For the following results, we use DAG-Bootstrap~\cite{agrawal2019abcd}, an approximate posterior method based on GIES causal discovery method~\cite{hauser2012characterization}. As GIES is not a differentiable method, once we compute the posterior, we transfer the posterior samples (the bootstraps) into JAX~\citep{jax} tensors to allow for the gradients to be computed with respect to the experiments.

\paragraph{Single-target synthetic graphs:} In this experiment, we test against synthetic graphs of 50 variables and batch size 5, where the graph is sampled from the class of Erdos-Renyi (a common benchmark in the literature~\cite{tigas2022interventions, toth2022active, scherrer2021learning}). In~\cref{fig:synthetic_results} \textbf{(A,B,C)} we summarize the results. We observe that our method performs significantly better than the baselines.
{
\small
\begin{table}[t!]
    \centering   
    \caption{Results of multi-target experiments on graphs of size 40 (30 seeds $\pm$ s.e.). Similarly to \cite{sussex2021near}, we are using posterior samples trained on observational data and re-weighting them with likelihoods.}
    \begin{tabular}{l|lll}
        \toprule
        \textbf{Method} & \textbf{$\mathbb{E}$SHD}~$\downarrow$ & \textbf{F1}~$\uparrow$ & \textbf{iMMD}~$\downarrow$ \\
        \midrule
        Random  &   $43.78{\color{black!50}\pm46.67}$ &   $0.91{\color{black!50}\pm0.08}$  & $0.16{\color{black!50}\pm0.07}$ \\
        SSGb    & $15.59{\color{black!50}\pm29.66}$ &   $0.97{\color{black!50}\pm0.05}$ & $0.10{\color{black!50}\pm0.06}$ \\
        \textbf{diffCBED}  & $\;\;0.44{\color{black!50}\pm0.21}$ &   $0.99{\color{black!50}\pm0.00}$ & $0.07{\color{black!50}\pm0.01}$ \\
        \bottomrule
    \end{tabular}
    \label{table:importance_weighted}
\end{table}
}%
\paragraph{$20$ nodes, unconstrained $(q \leq 20)$, batch size $B=2$:} In this experiment, we evaluate the performance of our method as compared to the baselines, on sparse graphs over several acquisitions.~\cref{fig:synthetic_results} \textbf{(D,E,F)} summarizes the results of this setting. We observe strong empirical performance as compared to all the baselines. Additional results are given in~\cref{app:synthetic_results}.
\subsection{Evaluation on Semi-Synthetic Data}
We evaluate our proposed design framework on semi-synthetic setting based on the DREAM gene networks~\citep{greenfield2010dream4}. In particular, we use the E. Coli gene interaction network, which is real-world inspired gene regulatory network of 10 variables, and simulate the mechanisms (and hence the data generation process). Then we run a random node with a fixed intervention value, a random node with a random value, SSGb finite sample and our algorithm. The results are presented in \cref{fig:semi_synthetic_results}. We find that our algorithm performs better than the baselines, indicating the potential of gradient-based approach to more realistic settings.
\section{Discussion}
\noindent{\textbf{Limitations:}} A primary limitation of our method is that it needs to estimate a posterior after every acquisition. While the proposed IWNMC estimator presents an interesting alternative, the designs are still non-adaptive. An interesting direction is to train a policy to design adaptive experiments~\citep{foster2021deep,greenewald2019sample}.

\noindent{\textbf{Conclusion:}} We present a gradient-based method for differentiable Bayesian optimal experimental design for causal discovery. Our method allows not only for single-target but also various multi-target (constrained and unconstrained) batch design of experiments. While prior work in relies on greedy approximations for the selection of a batch~\cite{agrawal2019abcd,tigas2022interventions} or black-box methods~\cite{toth2022active,tigas2022interventions} for optimizing over interventional states, our method utilizes gradient-based optimization procedures to simultaneously optimize for various design choices. Evaluation on different datasets suggests that our method is competitive with baselines.

\looseness=-1\paragraph{Acknowledgements:} We would like to thank Berzelius computing from NSC Sweden for providing computational resource for some of the experiments of the paper.
\bibliography{references}
\bibliographystyle{includes/icml2023}
\newpage
\onecolumn
\begin{appendices}
\section{Derivation of Importance Weighted Nested Monte Carlo Estimator}
\label{sec:app:iwnmc}
In this section, we derive the $\mathcal{U}_{\text{IWNMC}}$ (Eq.~\ref{eq:iwnmc}) estimator. We derive the estimator for a single design with an experiment denoted by $\xi$, parameters $\btheta$ and experimental outcome random variable $\mathbf{Y}$ and its instance $\mathbf{y}$. Since it is a static design, all the steps of the derivation hold if we replace $\xi$ with $\xi_{1:B}$, $\mathbf{Y}$ with $\mathbf{Y_{1:B}}$ and $\mathbf{y}$ with $\mathbf{y_{1:B}}$.
We begin from the variational NMC (VNMC) estimator, introduced by \citet{foster2019variational}
{\small
\begin{align}
    \mathcal{I}(\mathbf{Y};\mathbf{\Theta} \mid \xi) \le \mathcal{U}_\text{VNMC}(\xi) = 
    \E_{\substack{p(\btheta_{0} \mid h_{t-1})
    \\p(\mathbf{y} \mid \btheta_0, \xi)\\ q(\btheta_{1:L} \mid h_{t-1},\mathbf{y})}} \left[\log \frac{p(\mathbf{y}\mid \xi, \btheta_0)}{
    \frac{1}{L} \sum_{\ell=1}^L \frac{p(\mathbf{y} \mid \xi, \btheta_\ell)p(\btheta_\ell\mid h_{t-1} )}{q(\btheta_{1:L} \mid h_{t-1},\mathbf{y})}} \right].
\end{align}
}
This can be rewritten as
{\small
\begin{align}
    \mathcal{U}_\text{VNMC}(\xi) = 
    \E_{\substack{p(\btheta_{0} \mid h_{t-1})
    \\p(\mathbf{y} \mid \btheta_0, \xi)\\ q(\btheta_{1:L} \mid h_{t-1},\mathbf{y})}} \left[\log \frac{p(\mathbf{y}\mid \xi, \btheta_0)}{
    \frac{1}{L} \sum_{\ell=1}^L \frac{p(\mathbf{y} \mid \xi, \btheta_\ell)p(\btheta_\ell)p(h_{t-1} \mid \btheta_\ell )}{q(\btheta_{1:L} \mid h_{t-1},\mathbf{y})}} + \log p(h_{t-1}) \right]
\end{align}
}
and \citet{foster2019variational} observed that $\log p(h_{t-1})$ is a constant that does not depend on $\xi$ and so can be safely neglected when optimizing over designs.
If we take the original prior $p(\btheta_\ell)$ as our proposal distribution $q$, then we arrive at
{\small
\begin{align}
    \mathcal{U}_\text{VNMC-prior}(\xi) = 
    \E_{\substack{p(\btheta_{0} \mid h_{t-1})p(\btheta_{1:L})
    \\p(\mathbf{y} \mid \btheta_0, \xi)}} \left[\log \frac{p(\mathbf{y}\mid \xi, \btheta_0)}{
    \frac{1}{L} \sum_{\ell=1}^L p(\mathbf{y} \mid \xi, \btheta_\ell)p(h_{t-1} \mid \btheta_\ell )}  \right] + C
\end{align}
}
where $C=\log p(h_{t-1})$.
This allows us to sample contrastive samples from any distribution, but does not account for $\btheta_0$.
If we were to sample $\btheta_0$ from $p(\btheta_0)$, we can correct using an importance weight
{\small
\begin{align}
    \mathcal{U}_\text{VNMC-prior}(\xi) = 
    \E_{\substack{p(\btheta_{0:L})
    \\p(\mathbf{y} \mid \btheta_0, \xi)}} \left[\frac{p(\btheta_0 \mid h_{t-1})}{p(\btheta_0)}\log \frac{p(\mathbf{y}\mid \xi, \btheta_0)}{
    \frac{1}{L} \sum_{\ell=1}^L p(\mathbf{y} \mid \xi, \btheta_\ell)p(h_{t-1} \mid \btheta_\ell )}  \right] + C,
\end{align}
}
but unfortunately, this relies on knowing the density of the posterior or using the fact that $p(\btheta_0 \mid h_{t-1}) / p(\btheta_0) = p(h_{t-1} \mid \btheta_0) / p(h_{t-1})$, knowing the marginal likelihood of the data $h_{t-1}$. Neither of these is usually tractable.
Instead, we can use a self-normalized importance sampling approach, which amounts to estimating $p(h_{t-1})$ by a sum over $\theta_{0:L}$, giving the \emph{approximation} IWNMC:
{\small
\begin{align}
    \mathcal{U}_\text{IWNMC}(\xi) = 
    \E_{\substack{p(\btheta_{0:L})
    \\p(\mathbf{y} \mid \btheta_0, \xi)}} \left[\frac{p(h_{t-1} \mid \btheta_0)}{\frac{1}{L+1} \sum_{k=0}^L p(h_{t-1} \mid \btheta_k)}\log \frac{p(\mathbf{y}\mid \xi, \btheta_0)}{
    \frac{1}{L} \sum_{\ell=1}^L p(\mathbf{y} \mid \xi, \btheta_\ell)p(h_{t-1} \mid \btheta_\ell )}  \right] + C.
\end{align}
}
The form that is given in \eqref{eq:iwnmc} is obtained by first relabelling the $\btheta$ samples to start from 1
{\small
\begin{align}
    \mathcal{U}_\text{IWNMC}(\xi) = 
    \E_{\substack{p(\btheta_{1:L})
    \\p(\mathbf{y} \mid \btheta_1, \xi)}} \left[\frac{p(h_{t-1} \mid \btheta_1)}{\frac{1}{L} \sum_{k=1}^L p(h_{t-1} \mid \btheta_k)}\log \frac{p(\mathbf{y}\mid \xi, \btheta_1)}{
    \frac{1}{L-1} \sum_{\ell=2}^L p(\mathbf{y} \mid \xi, \btheta_\ell)p(h_{t-1} \mid \btheta_\ell )}  \right] + C,
\end{align}
}
noting that the role of $\btheta_1$ is arbitrary and can be replaced by any $m \in \{1,\dots,L\}$
{\small
\begin{align}
    \mathcal{U}_\text{IWNMC}(\xi) = 
    \E_{\substack{p(\btheta_{1:L})
    \\p(\mathbf{y} \mid \btheta_m, \xi)}} \left[\frac{p(h_{t-1} \mid \btheta_m)}{\frac{1}{L} \sum_{k=1}^L p(h_{t-1} \mid \btheta_k)}\log \frac{p(\mathbf{y}\mid \xi, \btheta_m)}{
    \frac{1}{L-1} \sum_{\ell\ne m}^L p(\mathbf{y} \mid \xi, \btheta_\ell)p(h_{t-1} \mid \btheta_\ell )}  \right] + C,
\end{align}
}
and finally taking the mean over $m$, noting that this does not change the expected value due to linearity
{\small
\begin{align}
    \mathcal{U}_\text{IWNMC}(\xi) = 
    \E_{\substack{p(\btheta_{1:L})
    \\p(\mathbf{y} \mid \btheta_m, \xi)}} \left[ \sum_{m=1}^L \frac{p(h_{t-1} \mid \btheta_m)}{\sum_{k=1}^L p(h_{t-1} \mid \btheta_k)}\log \frac{p(\mathbf{y}\mid \xi, \btheta_m)}{
    \frac{1}{L-1} \sum_{\ell\ne m}^L p(\mathbf{y} \mid \xi, \btheta_\ell)p(h_{t-1} \mid \btheta_\ell )}  \right] + C.
\end{align}
}
We finally drop the constant $C$ as it is independent of $\xi$ and take
\begin{equation}
    \omega_m = \frac{p(h_{t-1} \mid \btheta_m)}{\sum_{k=1}^L p(h_{t-1} \mid \btheta_k)}.
\end{equation}

\newpage
\section{Expected Information Gain for 6 Variables and Batch Size 2}
\label{app:eig2}

\begin{figure}[!h]
    \includegraphics[width=1\textwidth]{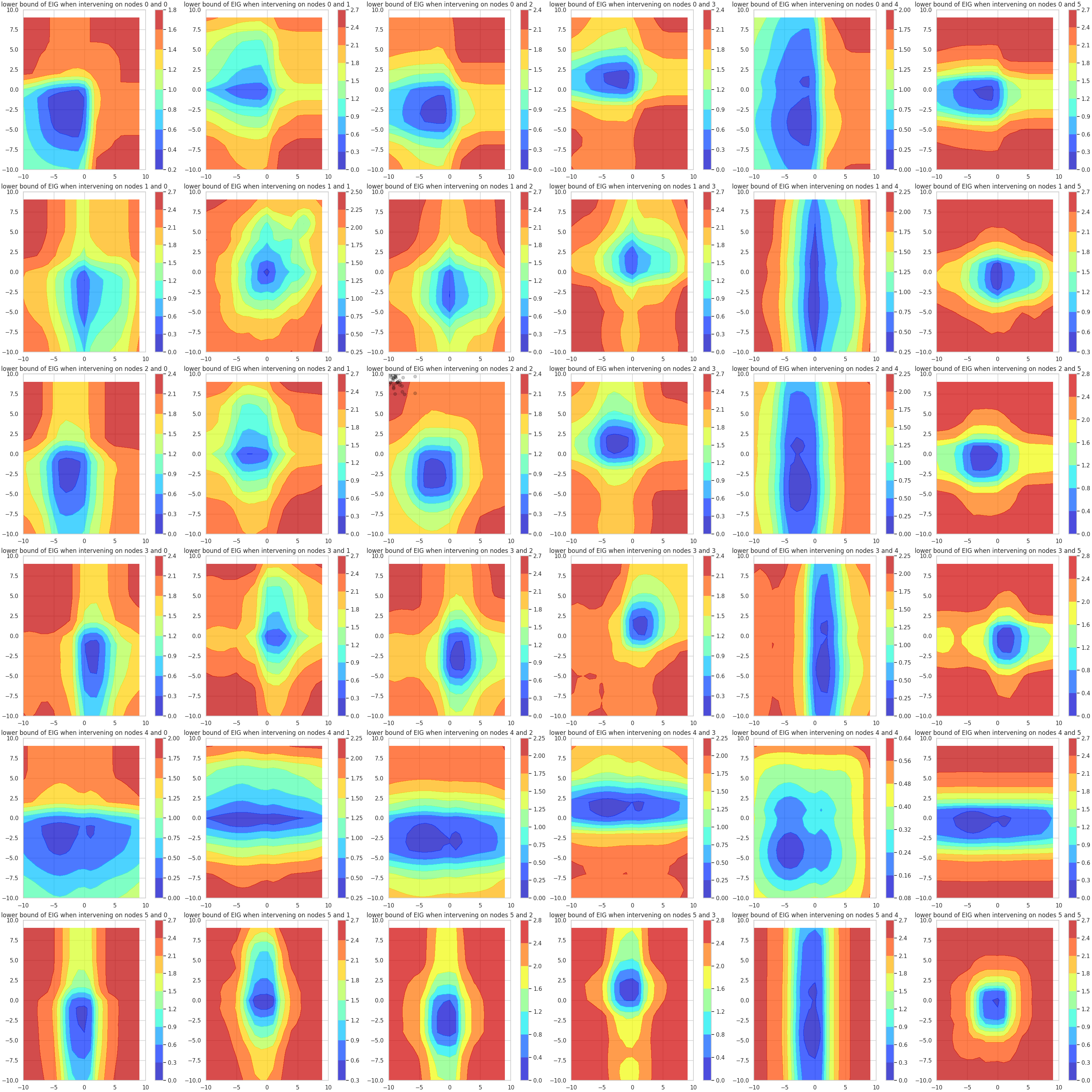}
    \caption{Here we visualize the Expected Information Gain of batch size two, on two nodes over different interventional states of the range $[-10, 10]$.}
    \label{fig:eigs_2d}
\end{figure}

\section{Metrics}
\label{app:metrics}.

\noindent{$\mathbb{E}$\textbf{-SHD}}: Defined as the \emph{expected structural hamming distance} between samples from the posterior model over graphs and the true graph $\mathbb{E}\text{-SHD} \coloneqq \mathbb{E}_{\mathbf{g} \sim p(\G \mid \mathcal{D})} \big[ \text{SHD}(\mathbf{g}, \tilde{\mathbf{g}}) \big]$

\noindent{\textbf{Expected edges F1}}: The expected F1 score of the binary classification task of predicting the presence/ absence of all edges. The expectation is taken over multiple posterior samples.

\noindent{\textbf{i-MMD}}: Interventional MMD is defined as MMD distance~\cite{gretton2012kernel} between the true interventional distribution and the interventional distribution induced by $\theta$ and $\mathbf{g}$ (posterior sample). We take an expectation over different posterior samples, interventional targets and interventional states. For the kernel choice, we use the median heuristic as described in~\cite{gretton2012kernel}.

\section{DAG Bootstrap}
\label{app:dag_bootstrap_settings}

The DAG bootstrap bootstraps observations and interventions to infer a different causal structure per bootstrap. We used GIES as the causal inference algorithm because of the adaptation of GES on interventional data as well. In our experiments, we used the pcalg R implementation \url{https://github.com/cran/pcalg/blob/master/R/gies.R} to discover 100 graphs. Each graph can be seen as a posterior sample from $p(\mathbf{G} \mid h_{t-1})$. For each of the sampled graphs $G_i$ we compute the appropriate $\theta_\text{MLE}$ under linear Gaussian assumption for the conditional distributions.

\section{Importance Weighted Nested Monte Carlo Full Results}\label{sec:IWNMC_full_results}

\begin{figure}[ht]
  \includegraphics[width=1\textwidth]{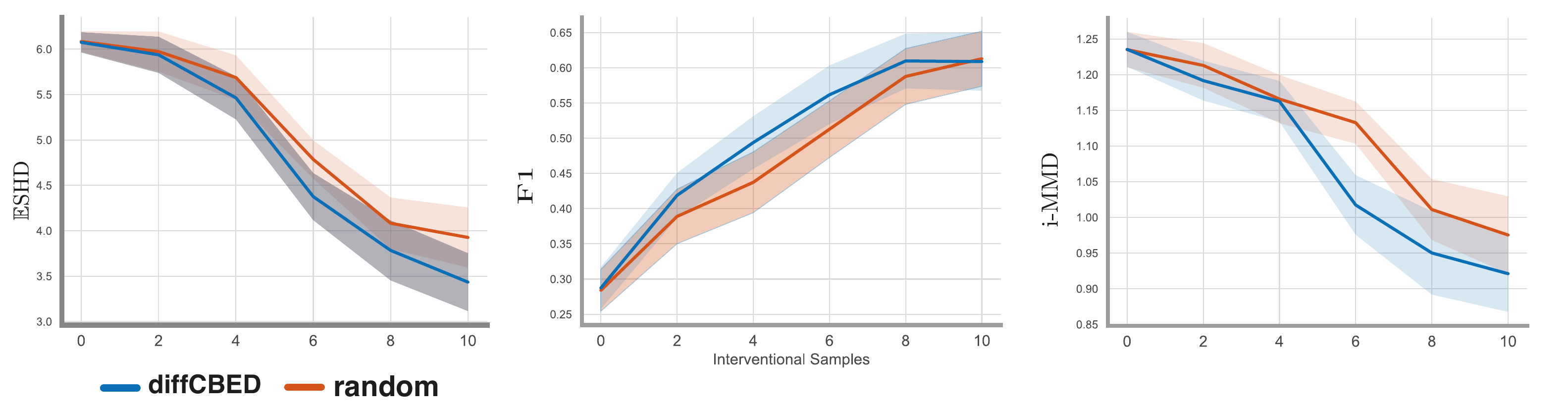} 
  \caption{Multi target-state design setting results for \erdos graphs with $d=5$ variables. Each experiment was run with 60 random seeds (shaded area represents 95\% CIs)}
  \label{fig:synthetic_results}  
\end{figure}

\begin{figure}[ht]
  \centering
  \includegraphics[width=.6\textwidth]{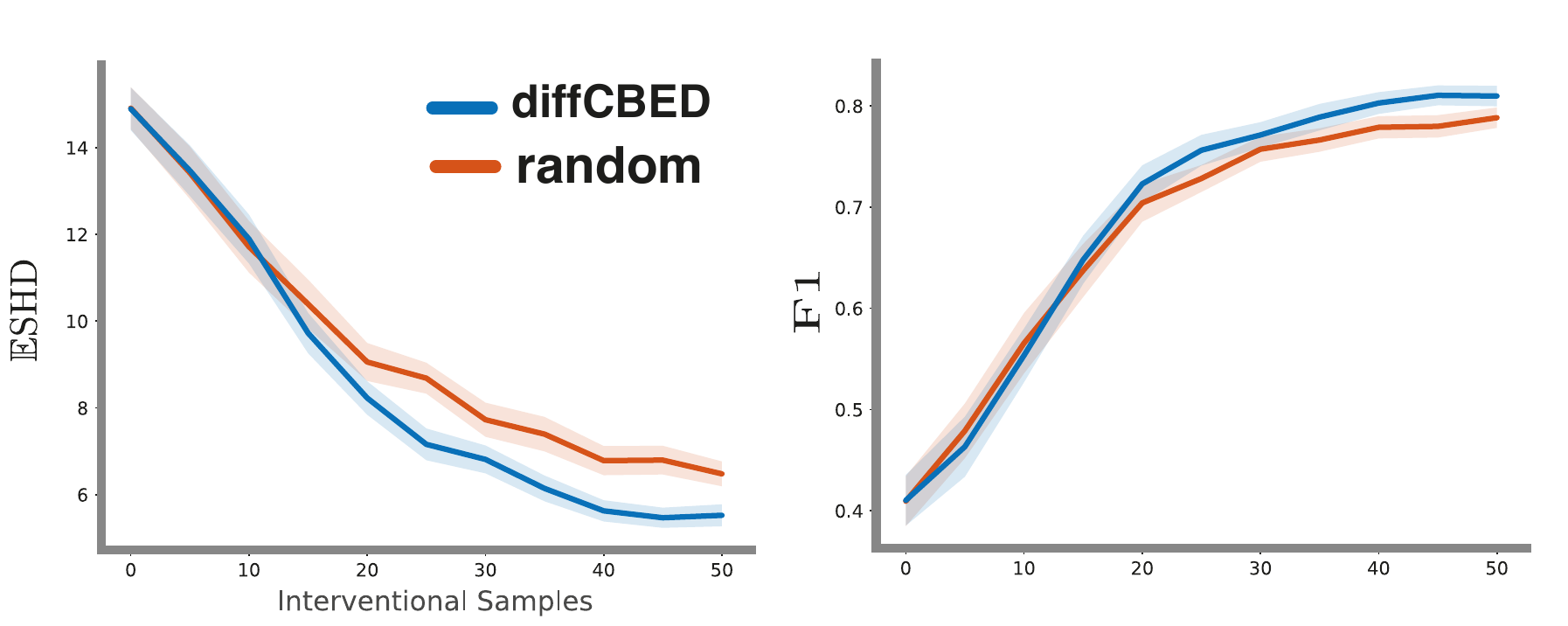} 
  \caption{Single target-state design setting results for \erdos graphs with $d=10$ variables. Each experiment was run with 60 random seeds (shaded area represents 95\% CIs)}
  \label{fig:10_iwnmc}  
\end{figure}

\newpage
\section{$20$ nodes, unconstrained $(q \leq 20)$, batch size $B=1$:}

\begin{figure}[ht]
  \includegraphics[width=1\textwidth]{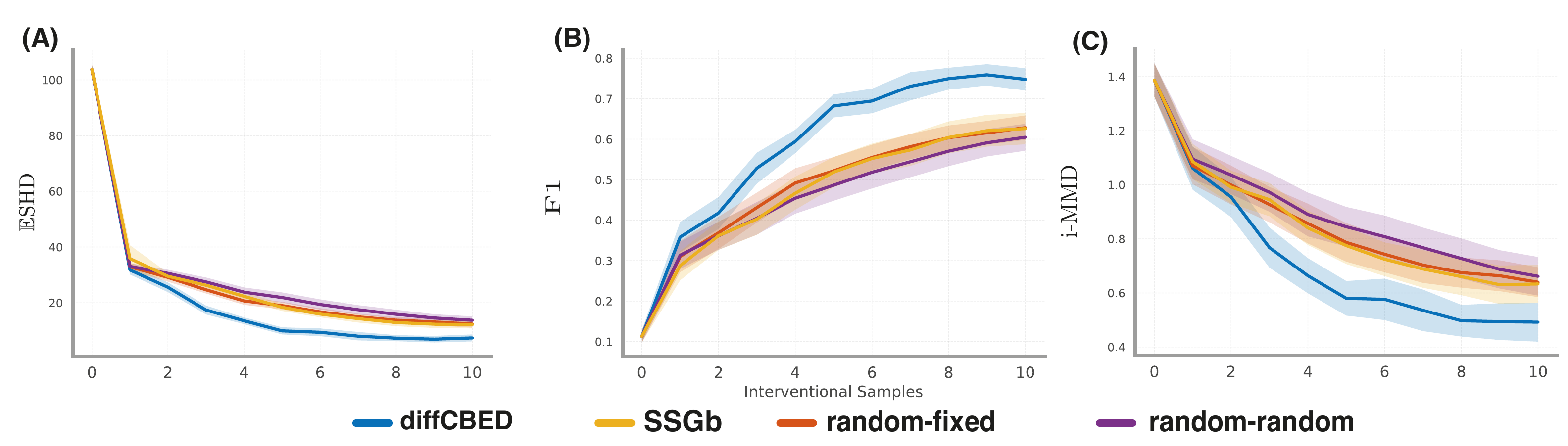} 
  \caption{Multi target-state design setting results for \erdos graphs with $d=50$ variables. Each experiment was run with 30 random seeds (the shaded area represents 95\% CIs). We observe that for batch size $1$, the difference between the methods becomes more significant.}
  \label{fig:synthetic_results}  
\end{figure}

\section{Datasets and Experiment Details}

\subsection{Synthetic Graphs Experiments}
\label{app:synthetic_results}

In the synthetic data experiments, we focus on Erdős-Rényi 
 graph model.
We used \texttt{networkx}\footnote{\url{https://networkx.org/documentation/networkx-1.10/reference/generated/networkx.generators.random_graphs.fast_gnp_random_graph.html}} and method \texttt{fast\_gnp\_random\_graph}~\citep{batagelj2005efficient} to generate graphs based on the Erdős-Rényi model. We set the expected number of edges per vertex to 1.

\section{Optimizer Settings}
\begin{table}[h!]
\caption{Table indicating the hyperparameters and optimizer settings for different experimental results.}
\begin{adjustbox}{width=1\textwidth}
\begin{tabular}{lcccc}
\hline
\multicolumn{5}{c}{\textbf{Optimization settings}}                                                                                                                                                                                                                                                                                                                                                               \\ \hline
\multicolumn{1}{l|}{}                                           & \multicolumn{1}{c|}{\begin{tabular}[c]{@{}c@{}}Single Target\\ NMC\end{tabular}} & \multicolumn{1}{c|}{\begin{tabular}[c]{@{}c@{}}Multi-Target\\ NMC\end{tabular}} & \multicolumn{1}{c|}{\begin{tabular}[c]{@{}c@{}}Multi-Target\\ IWNMC with prior\end{tabular}} & \begin{tabular}[c]{@{}c@{}}Multi-Target\\ IWNMC with proposal\end{tabular} \\ \hline
\multicolumn{1}{l|}{$L$}  & $30$ & $30$ & $1000$ & $60$ \\
\multicolumn{1}{l|}{Number of outer DAGs $N_o$} & $30$ & $30$ & $1000$ & $60$ \\
\multicolumn{1}{l|}{Batch Size} & $5$ & $2$ & $2$ & $2$ \\
\multicolumn{1}{l|}{Relaxation temperature} & $5 \rightarrow .5$ & $5 \rightarrow .5$  & $0.1$ & $5 \rightarrow .5$ \\
\multicolumn{1}{l|}{Optimizer} & Adam & Adam & Adam & Adam \\
\multicolumn{1}{l|}{Learning rate of optimizer} & $0.1$ & $0.1$ & $0.01$ & $0.1$ \\
\multicolumn{1}{l|}{Number of starting samples (observational)} & $60$ & $60$ & $2$ & $800$\\
\multicolumn{1}{l|}{Number of batches} & $10$ & $10$ & $5$ & $1$ \\
\multicolumn{1}{l|}{Number of DAG Bootstraps} & $30$ & $30$ & - & $60$ \\
\multicolumn{1}{l|}{Number of training steps per batch} & $100$ &  $100$ & $100$ & $100$\\
\end{tabular}
\end{adjustbox}
\end{table}
\section{Relaxed Distribution Temperature Sensitivity Analysis}
\begin{figure}[!h]
    \includegraphics[width=1\textwidth]{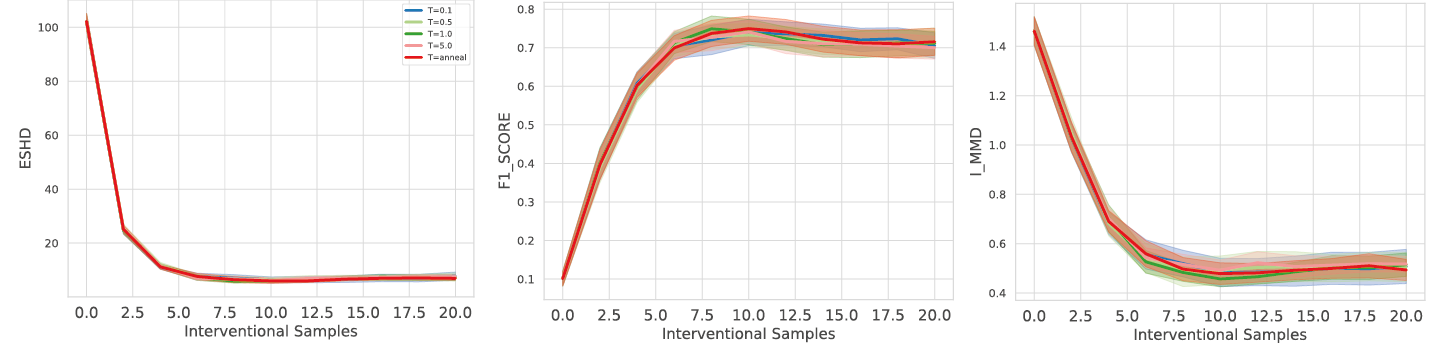}
    \caption{Performance of our proposed design strategy for different temperatures of the relaxed distribution for the unconstrained multi-target setting $(d=20)$. We find that our method is fairly robust.}
    \label{fig:temp_senitivity}
\end{figure}

We perform ablation where we empirically test the sensitivity to temperature hyperparameter for the relaxed distribution. Results are presented in \cref{fig:temp_senitivity} for the unconstrained multi-target setting ($d=20$). We find that our approach is fairly robust to different temperatures. Note that for the reported results, we anneal the temperature during the course of training and hence a single choice of temperature is not necessary.
\end{appendices}
\end{document}

%% file: macros.tex
\newcommand{\btheta}{\boldsymbol{\theta}}
\newcommand{\bTheta}{\boldsymbol{\Theta}}
\newcommand{\bgamma}{\boldsymbol{\gamma}}
\newcommand{\bg}{\mathbf{g}}

\newcommand{\G}{\mathcal{G}}

\newcommand{\erdos}{Erdős–Rényi~\cite{erdHos1959random}~}

\newcommand{\E}{\mathop{\mathbb{E}}}

\DeclareMathOperator*{\argmax}{arg\,max}

\newcommand{\bphi}{\boldsymbol{\phi}}

\newcommand{\design}{\{(I, S^I)\}}